\documentclass[journal]{IEEEtran}

\usepackage[hyphens]{url}
\newcommand{\href}[2]{#1}
\usepackage{cite}
\ifCLASSINFOpdf
  \usepackage[pdftex]{graphicx}
  \DeclareGraphicsExtensions{.pdf}
\else
\fi
\usepackage[cmex10]{amsmath}
\usepackage{eufrak}
\usepackage{algorithm}
\usepackage{algpseudocode}
\usepackage{multirow}
\usepackage[caption=false,font=footnotesize]{subfig}

\begin{document}
\title{Appearance-Based Loop Closure Detection for Online Large-Scale and Long-Term Operation}

\author{Mathieu~Labb\'e,~\IEEEmembership{Student Member,~IEEE,}
        Fran{\c c}ois~Michaud,~\IEEEmembership{Member,~IEEE}
\thanks{
Manuscript received April 23, 2012; revised October 2, 2012; accepted January 14, 2013. This paper was recommended for publication by Associate Editor P. Jensfelt and Editor D. Fox upon evaluation of the reviewers’ comments.
This work was supported in part by the Natural Sciences and Engineering Research Council of Canada, the Canadian Foundation for Innovation and the Canada Research Chair program.}
\thanks{M. Labb\'e and F. Michaud are with the Department of Electrical and Computer Engineering, Universit\'e de Sherbrooke, Sherbrooke, QC, CA J1K 2R1 (e-mail:\{mathieu.m.labbe,francois.michaud\}@usherbrooke.ca).}
\thanks{Color versions of one or more of the figures in this paper are available online at http://ieeexplore.ieee.org.}
\thanks{Digital Object Identifier 10.1109/TRO.2013.2242375}
}

\maketitle

\begin{abstract}
In appearance-based localization and mapping, loop closure detection is the process used to determinate if the current observation comes from a previously visited location or a new one. As the size of the internal map increases, so does the time required to compare new observations with all stored locations, eventually limiting online processing. This paper presents an online loop closure detection approach for large-scale and long-term operation. The approach is based on a memory management method, which limits the number of locations used for loop closure detection so that the computation time remains under real-time constraints. The idea consists of keeping the most recent and frequently observed locations in a Working Memory (WM) used for loop closure detection, and transferring the others into a Long-Term Memory (LTM). When a match is found between the current location and one stored in WM, associated locations stored in LTM can be updated and remembered for additional loop closure detections.
Results demonstrate the approach's adaptability and scalability using ten standard data sets from other appearance-based loop closure approaches, one custom data set using real images taken over a 2 km loop of our university campus, and one custom data set (7 hours) using virtual images from the racing video game ``Need for Speed: Most Wanted''. 
\end{abstract}


\section{Introduction}
\IEEEPARstart{A}{utonomous} robots operating in real life settings must be able to navigate in large, unstructured, dynamic and unknown spaces. Simultaneous localization and mapping (SLAM) \cite{thrun05} is the capability required by robots to build and update a map of their operating environment and to localize themselves in it. A key feature in SLAM is to recognize previously visited locations. This process is also known as loop closure detection, referring to the fact that coming back to a previously visited location makes it possible to associate this location with another one recently visited. 

For most of the probabilistic SLAM approaches \cite{montemerlo2003fastslam, Bosse04, Eliazar04, Estrada05, Folkesson07, Clemente07, Grisetti07b, Blanco08, callmer2008tree, Paz08, Pinies08, schleicher2010real}, loop closure detection is done locally, i.e., matches are found between new observations and a limited region of the map, determined by the uncertainty associated with the robot's position. Such approaches can be processed under real-time contraints at 30 Hz \cite{davison2007monoslam} as long as the estimated position is valid, which cannot be guaranteed in real world situations \cite{Newman06}. As an exclusive or complementary alternative, appearance-based loop closure detection approaches \cite{Newman06, Cummins08a, Angeli08b, botterill2011bag, konolige2010view} generally detect a loop closure by comparing a new location with all previously visited locations, independently of the estimated position of the robot. If no match is found, then a new location is added to the map. However, a robot operating in large areas for a long period of time will ultimately build a very large map, and the amount of time required to process new observations increases with the number of locations in the map. If computation time becomes larger than the acquisition time, a delay is introduced, making updating and processing the map difficult to achieve online. 

Our interest lies in developing an online appearance-based loop closure detection approach that can deal with large-scale and long-term operation. 
Our approach dynamically manages the locations used to detect loop closures, in order to limit the time required to search through previously visited locations. This paper describes our memory management approach to accomplish appearance-based loop closure detection, in a Bayesian framework, with real-time constraints for large-scale and long-term operation. Processing time, i.e., the time required to process an acquired image, is the criterion used to limit the number of locations kept in the robot's Working Memory (WM). To identify the locations to keep in WM, the solution studied in this paper consists of keeping the most recent and frequently observed locations in WM, and transferring the others into Long-Term Memory (LTM). When a match is found between the current location and one stored in WM, associated locations stored in LTM can be remembered and updated. This idea is inspired from observations made by psychologists \cite{atkinson1968human, baddeley1997human} that people remembers more the areas where they spent most of their time, compared to those where they spent less time. By following this heuristic, the compromise made between search time and space is therefore driven by the environment and the experiences of the robot. 

Because our memory management mechanism is made to ensure satisfaction of real-time constraints for online processing (in the sense that the time required to process new observations remains lower or equal to the time to acquire them, as in \cite{Estrada05, schleicher2010real, davison2007monoslam, galvez2011real}), independently of the scale of the mapped environment, our approach is named Real-Time Appearance-Based Mapping (RTAB-Map\footnote{Open source software available at \href{http://rtabmap.googlecode.com}{http://rtabmap.googlecode.com}.}). An earlier version of RTAB-Map has been presented in \cite{labbe11memory}. This paper presents in more details the improved version tested with a much wider set of conditions, and is organized as follows. Section \ref{sec:related_work} reviews appearance-based loop closure detection approaches. Section \ref{sec:system_description} describes RTAB-Map. Section \ref{sec:results} presents experimental results, and Section \ref{sec:discussion} presents limitations and possible extensions to our approach. 

\section{Related work}
\label{sec:related_work}
For global loop closure detection, vision is the sense generally used to derive observations from the environment because of the distinctiveness of features extracted from the environment \cite{Milford08, Newman05, Tapus08}, although successful large-scale mapping using laser range finder data is possible \cite{Bosse07}. 
For vision-based mapping, the bag-of-words \cite{sivic2003video} approach is commonly used \cite{Nister06, Cummins08a, Angeli08c, botterill2011bag, konolige2010view} and has shown to perform online loop closure detection for paths of up to 1000 km \cite{cummins2009highly}.
The bag-of-words approach consists in representing each image by visual words taken from a vocabulary.
The visual words are usually made from local feature descriptors, such as Scale-Invariant Feature Transform (SIFT) \cite{Lowe04}. 
These features have high dimensionality, making it important to quantize them into a vocabulary for fast comparison instead of making direct comparisons between features. Popular quantization methods are Vocabulary Tree \cite{Nister06}, Approximate K-Means \cite{philbin2007object} or K-D Tree \cite{Lowe04}. By linking each word to related images, comparisons can be done efficiently over large data sets, as with the Term Frequency-Inverse Document Frequency (TF-IDF) approach \cite{sivic2003video}. 

The vocabulary can be constructed offline (using a training data set) or incrementally constructed online, although the first approach is usually preferred for online processing in large-scale environments. 
However, even if the image comparison using a pre-trained vocabulary (as in FAB-MAP 2.0 \cite{cummins2009highly}) is fast, real-time constraints satisfaction is effectively limited by the maximum size of the mapped environment.
The number of comparisons can be decreased by considering only a selection of previously acquired images (referred to as key images) for the matching process, while keeping detection performance nearly the same compared to using all images \cite{booij2009efficient}. Nevertheless, processing time for each image acquired still increases with the number of key images. In \cite{ranganathan2010pliss}, a particle filter is used to detect transition between sets of locations referred to categories, but again, the cost of updating the place distribution increases with the number of categories.

Even if the robot is always revisiting the same locations in a closed environment, perceptual aliasing, changes that can occur in dynamic environments or the lack of discriminative information may affect the ability to recognize previously visited locations. This leads to the addition of new locations in the map and consequently influences the satisfaction of real-time constraints \cite{glover10}.
To limit the growth of images to match, pruning \cite{milford2010persistent} or clustering \cite{konolige2009towards} methods can be used to consolidate portions of the map that exceed a spatial density threshold. 
This limits growth over time, but not in relation to the size of the explored environment.

Finally, memory management approaches have been used in robot localization to increase recognition performance in dynamical environments \cite{Dayoub08} or to limit memory used\cite{pronobis2010more}. In contrast, our memory management is used for online localization and mapping, where new locations are dynamically added over time.

\section{Online Appearance-Based Mapping}
\label{sec:system_description}
The objective of our work is to provide an appearance-based localization and mapping solution independent of time and size, to achieve online loop closure detection for long-term operation in large environments. The idea resides in only using a limited number of locations for loop closure detection so that real-time constraints can be satisfied, while still gain access to locations of the entire map whenever necessary. When the number of locations in the map makes processing time for finding matches greater than a time threshold, our approach transfers locations less likely to cause loop closure detection from the robot's WM to LTM, so that they do not take part in the detection of loop closures. However, if a loop closure is detected, neighbor locations can be retrieved and brought back into WM to be considered in future loop closure detections. 

\begin{figure}[!t] 
\centering 
\includegraphics[width= 2.3in]{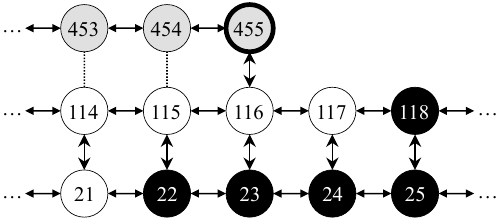} 
\caption[Graph]{Graph representation of locations. Vertical arrows are loop closure links and horizontal arrows are neighbor links. Dotted links show not detected loop closures. Black locations are those in LTM, white ones are in WM and gray ones are in STM. Node 455 is the current acquired location.} 
\label{fig_graph} 
\end{figure}

As an illustrative example used throughout this paper, \figurename \ref{fig_graph} shows a graph representation of locations after three traversals of the same region. Each location is represented by an image signature, a time index (or age) and a weight, and locations are linked together in a graph by neighbor or loop closure links. These links represent locations near in time or in space, respectively. 

Locations in LTM are not used for loop closure detection. Therefore, it is important to choose carefully which locations to transfer to LTM. A naive approach is to use a first-in first-out (FIFO) policy, pruning the oldest locations from the map to respect real-time constraints. However, this sets a maximum sequence of locations that can be memorized when exploring an environment: if the processing time reaches the time threshold before loop closures can be detected, pruning the older locations will make it impossible to find a match. As an alternative, a location could be randomly picked, but it is preferable to keep in WM the locations that are more susceptible to be revisited.
As explained in the introduction, the idea studied in this paper is based on the working hypothesis that locations seen more frequently than others are more likely to cause loop closure detections. 
Therefore, the number of time a location has been consecutively viewed is used to set its weight. When a transfer from WM to LTM is necessary, the location with the lowest weight is selected. If many locations have the same lowest weight, the oldest one is transferred. 

\begin{figure}[!t] 
\centering 
\includegraphics[width= 2.2in]{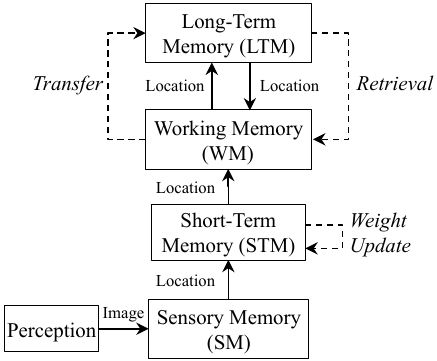} 
\caption[Memory management model]{RTAB-Map memory management model.} 
\label{fig_summary} 
\end{figure}

\figurename \ref{fig_summary} illustrates RTAB-Map memory management model. 
The Perception module acquires an image and sends it to Sensory Memory (SM). SM evaluates the image signature to reduce data dimensionality and to extract useful features for loop closure detection. Then SM creates a new location with the image signature and sends it to Short-Term Memory (STM).
STM updates recently created locations through a process referred to as Weight Update. 
If Weight Update considers that the new location is similar to the last one in STM, it merges them into the new one, and it increases the weight of the new location. 

STM is used to observe similarities through time between consecutive images for weight updates, while the role of the WM is to detect loop closures between locations in space.
Similar to \cite{Angeli08b}, RTAB-Map does not use locations in STM to avoid loop closures on locations that have just been visited (because most of the time, the last location frequently looks similar to the most recent ones). 
The STM size $T_{\mathrm{STM}}$ is set based on the robot velocity and the rate at which the locations are acquired. When the number of locations in STM reaches $T_{\mathrm{STM}}$, the oldest location in STM is moved into WM.
RTAB-Map evaluates loop closure probabilities with a discrete Bayesian filter by comparing the new location with the ones in WM. 
A loop closure is detected and locations are linked together when a high loop closure probability is found between a new and an old location in WM.
Two steps are then key in ensuring that locations more susceptible to cause future loop closure detections are in WM while keeping the WM size under an online limit tractable by the Bayesian filter.
The first step is called Retrieval: neighbor locations of the highest loop closure probability, for those which are not in WM, are brought back from LTM into WM, increasing the probability of identifying loop closures with future nearby locations.
The second step is called Transfer: if the processing time for loop closure detection is greater than the time threshold $T_{\mathrm{time}}$, the oldest of the least viewed locations (i.e., oldest locations with smallest weights) are transferred to LTM. 
The number of transferred locations depends on the number of locations added to WM during the current cycle. 

Algorithm \ref{alg:mainLoop} illustrates the overall loop closure detection process, explained in details in the following sections. 

\begin{algorithm}\small
\begin{algorithmic}[1]
\State $time\gets \Call{TimeNow}{ }$ \Comment $\Call{TimeNow}{ }$ returns current time
\State $I_t\gets$ acquired image
\State $L_t\gets$ \Call{LocationCreation}{$I_t$}
\If{$z_t$ (of $L_t$) is a bad signature (using $T_{\mathrm{bad}}$)}
\State Delete $L_t$
\Else
\State Insert $L_t$ into STM, adding a neighbor link with $L_{t-1}$
\State Weight Update of $L_t$ in STM (using $T_{\mathrm{similarity}}$)
\If{STM's size reached its limit ($T_{\mathrm{STM}}$)}
\State Move oldest location of STM to WM
\EndIf
\State $\boldsymbol{p}(S_t|L^t)\gets$Bayesian Filter Update in WM with $L_t$
\State Loop Closure Hypothesis Selection ($S_t=i$)
\If{$S_t=i$ is accepted (using $T_{\mathrm{loop}}$)}
\State Add loop closure link between $L_t$ and $L_i$
\EndIf
\State Join $trash$'s thread \Comment Thread started in $\Call{Transfer}{ }$
\State \Call{Retrieval}{$L_i$} \Comment LTM $\rightarrow$ WM
\State $pTime\gets \Call{TimeNow}{ } - time$ \Comment Processing time
\If{$pTime > T_{\mathrm{time}}$}
\State $\Call{Transfer}{ }$ \Comment WM $\rightarrow$ LTM
\EndIf
\EndIf
\end{algorithmic}
\caption{RTAB-Map} 
\label{alg:mainLoop} 
\end{algorithm}

\subsection{Location Creation}
\label{sec:location}

The bag-of-words approach \cite{sivic2003video} is used to create signature $z_t$ of an image acquired at time $t$.  
An image signature is represented by a set of visual words contained in a visual vocabulary incrementally constructed online. We chose to use an incremental rather than a pre-trained vocabulary to avoid having to go through a training step for the targeted environment. 

Using OpenCV \cite{opencv_library}, Speeded-Up Robust Features (SURF) \cite{Bay08} are extracted from the image to derive visual words. Each visual word of the vocabulary refers to a single SURF feature's descriptor (a vector of 64 dimensions).
Each SURF feature has a strength referred to as feature response.
The feature response is used to select the most prominent features in the image. To avoid bad features, only those over a feature response of $T_{\mathrm {response}}$ are extracted.
A maximum of $T_{\mathrm {maxFeatures}}$ SURF features with the highest feature response are kept to have nearly the same number of words across the images. 
If few SURF features are extracted (under a ratio $T_{\mathrm{bad}}$ of the average features per image), the signature is considered to be a bad signature and is not processed for loop closure detection. This happens when an image does not present discriminative features, such as a white wall in an indoor scene. 

For good signatures, to find matches with words already in the vocabulary (a process referred to as the quantization step), SURF features are compared using the distance ratio between the nearest and the second-nearest neighbor (called nearest neighbor distance ratio, NNDR). 
As in \cite{Lowe04}, two features are considered to be represented by the same word if the distance with the nearest neighbor is less than $T_{\mathrm {NNDR}}$ times the distance to the second-nearest neighbor. 
Because of the high dimensionality of SURF descriptors, a randomized forest of four kd-trees (using FLANN \cite{muja_flann_2009}) is used. This structure increases efficiency of the nearest-neighbor search when matching descriptors from a new signature with the ones associated to each word in the vocabulary (each leaf of the kd-trees corresponds to a word in the vocabulary). The randomized kd-trees approach was chosen over the hierarchical k-means approach because of its lower tree-build time \cite{muja_flann_2009}: FLANN does not provide an interface for incremental changes to its search indexes (such as randomized kd-trees or hierarchical k-means tree), so they need to be reconstructed online at each iteration as the vocabulary is modified.
The kd-trees are built from all the SURF descriptors of the words contained in the vocabulary. Then, each descriptor extracted from the image is quantized by finding the two nearest neighbors in the kd-trees. 
For each extracted feature, when $T_{\mathrm {NNDR}}$ criterion is not satisfied, a new word is created with the feature's descriptor. The new word is then added to the vocabulary and $z_t$. If the match is accepted with the nearest descriptor, its corresponding word of the vocabulary is added to $z_t$. 

A location $L_t$ is then created with signature $z_t$ and time index $t$; its weight initialized to 0 and a bidirectional link in the graph with $L_{t-1}$. The summary of the location creation procedure is shown in Algorithm \ref{alg:createLocation}. 

\begin{algorithm}\small
\begin{algorithmic}[1]
\Procedure{LocationCreation}{$I$}
\State $f\gets$ detect a maximum of $T_{\mathrm{maxFeatures}}$ SURF features from
\Statex\hspace{\algorithmicindent}image $I$ with SURF feature response over $T_{\mathrm{response}}$
\State $d\gets$ extract SURF descriptors from $I$ with features $f$
\State Prepare nearest-neighbor index (build kd-trees)
\State $z\gets$ quantize descriptors $d$ to vocabulary (using kd-trees and 
\Statex\hspace{\algorithmicindent}$T_{\mathrm{NNDR}}$)
\State $L\gets$ create location with signature $z$ and weight 0
\State \Return $L$
\EndProcedure
\end{algorithmic}
\caption{Create location $L$ with image $I$} 
\label{alg:createLocation} 
\end{algorithm}

\subsection{Weight Update}
\label{sec_memory_rehearsal}

To update the weight of the acquired location, $L_t$ is compared to the last one in STM, and similarity $\mathfrak{s}$ is evaluated using (\ref{eq:similarity}) :

\begin{equation}	
	\label{eq:similarity}
	\mathfrak{s}(z_t, z_c)=\left\{\begin{matrix}
N_{\mathrm{pair}}/N_{z_t}, & \textrm{if } N_{z_t}\geq N_{z_c} \\ 
N_{\mathrm{pair}}/N_{z_c}, & \textrm{if } N_{z_t} < N_{z_c}
\end{matrix}\right. 
\end{equation}

\noindent where $N_{\mathrm{pair}}$ is the number of matched word pairs between the compared location signatures, and where $N_{z_t}$ and $N_{z_c}$ are the total number of words of signature $z_t$ and the compared signature $z_c$ respectively. 
If $\mathfrak{s}(z_t, z_c)$ is higher than a fixed similarity threshold $T_{\mathrm {similarity}}$ (ratio between 0 and 1), the compared location $L_c$ is merged into $L_t$. Only the words from $z_c$ are kept in the merged signature, and the newly added words from $z_t$ are removed from the vocabulary: $z_t$ is cleared and $z_c$ is copied into $z_t$.
The reason why $z_t$ is cleared is that it is easier to remove the new words of $z_t$ from the vocabulary because their descriptors are not yet indexed in the kd-trees (as explained in Section \ref{sec:location}). Empirically, we found that similar performances are observed when only the words of $z_t$ are kept or that both are combined, using a different $T_{\mathrm{similarity}}$. In all cases, words found in both $z_c$ and $z_t$ are kept in the merged signature, the others are generally less discriminative.
To complete the merging process, the weight of $L_t$ is increased by the weight of $L_c$ plus one, the neighbor and loop closure links of $L_c$ are redirected to $L_t$, and $L_c$ is deleted from STM.

\subsection{Bayesian Filter Update}
\label{sec:bayes}

The role of the discrete Bayesian filter is to keep track of loop closure hypotheses by estimating the probability that the current location $L_t$ matches one of an already visited location stored in the WM. 
Let $S_t$ be a random variable representing the states of all loop closure hypotheses at time $t$. 
$S_t = i$ is the probability that $L_t$ closes a loop with a past location $L_i$, thus detecting that $L_t$ and $L_i$ represent the same location. $S_t = -1$ is the probability that $L_t$ is a new location.
The filter estimates the full posterior probability $\boldsymbol {p}(S_t|L^t)$ for all $i=-1,...,t_{n}$, where $t_{n}$ is the time index associated with the newest location in WM, expressed as follows \cite{Angeli08c}: 

\begin{equation}	
	\label{eq:bayes}
\boldsymbol{p}(S_t|L^t) = \eta 
\underbrace{
\boldsymbol{p}(L_t|S_t) 
}_\textrm{Observation}
\underbrace{
\displaystyle\sum\limits_{i=-1}^{t_n} 
\underbrace{
\boldsymbol{p}(S_t|S_{t-1}=i) 
}_\textrm{Transition}
p(S_{t-1}=i|L^{t-1})
}_\textrm{Belief}
\end{equation}

\noindent where $\eta$ is a normalization term and $L^{t} = L_{-1},...,L_{t}$. Note that the sequence of locations $L^t$ includes only the locations contained in WM and STM. Therefore, $L^t$ changes over time as new locations are created or when some locations are retrieved from LTM or transferred to LTM, in contrast to the classical Bayesian filtering where such sequences are fixed.

The observation model $\boldsymbol{p}(L_t|S_t)$ is evaluated using a likelihood function $\mathfrak{L}(S_t|L_t)$ : the current location $L_t$ is compared using (\ref{eq:similarity}) with locations corresponding to each loop closure state $S_t=j$ where $j=0,..,t_n$, giving a score $s_j=\mathfrak{s}(z_t,z_j)$.
The difference between each score $s_j$ and the standard deviation $\sigma$ is then normalized by the mean $\mu$ of all non-null scores, as in (\ref{eq:normalizedscore}) \cite{Angeli08c}  :
\begin{equation}
	\label{eq:normalizedscore}
p(L_t|S_t=j) = \mathfrak{L}(S_t=j|L_t)= \left\{\begin{matrix}
\frac{s_j-\sigma}{\mu}, & \textrm{if } s_j\geq \mu+\sigma \\ 
1, & \textrm{otherwise.}
\end{matrix}\right. 
\end{equation}
\noindent For the new location probability $S_t=-1$, the likelihood is evaluated using (\ref{eq:virtualPlaceLikelihood}) :
\begin{equation}	
	\label{eq:virtualPlaceLikelihood}
	p(L_t|S_t=-1) = \mathfrak{L}(S_t=-1|L_t) = \frac{\mu}{\sigma} + 1
\end{equation}
\noindent where the score is relative to $\mu$ on $\sigma$ ratio. 
If $\mathfrak{L}(S_t=-1|L_t)$ is high (i.e., $L_t$ is not similar to a particular location in WM, as $\sigma<\mu$), then $L_t$ is more likely to be a new location. 

The transition model $\boldsymbol {p}(S_t|S_{t-1}=i)$ is used to predict the distribution of $S_t$, given each state of the distribution $S_{t-1}$ in accordance with the robot's motion between $t$ and $t-1$. Combined with $p(S_{t-1}=i|L^{t-1})$ (i.e., the recursive part of the filter), this constitutes the belief of the next loop closure. 
The transition model is expressed as in \cite{Angeli08c}:
\begin{enumerate}
  \item $p(S_t=-1|S_{t-1}=-1)=0.9$, the probability of a new location event at time $t$ given that no loop closure occurred at time $t-1$.
  \item $p(S_t=i|S_{t-1}=-1)=0.1/N_{\mathrm {WM}}$ with $i\in[0;t_n]$, the probability of a loop closure event at time $t$ given that no loop closure occurred at $t-1$. $N_{\mathrm{WM}}$ is the number of locations in WM of the current iteration.
  \item $p(S_t=-1|S_{t-1}=j)=0.1$ with $j\in[0;t_n]$, the probability of a new location event at time $t$ given that a loop closure occurred at time $t-1$ with $j$.
  \item $p(S_t=i|S_{t-1}=j)$ with $i,j\in[0;t_n]$, the probability of a loop closure event at time $t$ given that a loop closure occurred at time $t-1$ on a neighbor location. The probability is defined as a discretized Gaussian curve ($\sigma=1.6$) centered on $j$ and where values are non-null for a neighborhood range of sixteen neighbors (for $i=j-16,...,j+16$).
Within the graph, a location can have more than two adjacent neighbors (if it has a loop closure link) or some of them are not in WM (because they were transferred to LTM). The Gaussian's values are set recursively by starting from $i=j$ to the end of the neighborhood range (i.e., sixteen), then $p(S_t>=0|S_{t-1}=j)$ is normalized to sum $0.9$. 
\end{enumerate}

\subsection{Loop Closure Hypothesis Selection}
\label{sec:loop_selection}

When $\boldsymbol {p}(S_t|L^t)$ has been updated and normalized, the highest loop closure hypothesis $S_t = i$ of $\boldsymbol {p}(S_t|L^t)$ is accepted if the new location hypothesis $p(S_t=-1|L^t)$ is lower than the loop closure threshold $T_{\mathrm{loop}}$ (set between 0 and 1).
When a loop closure hypothesis is accepted, $L_t$ is linked with the old location $L_i$: the weight of $L_t$ is increased by the one of $L_i$, the weight of $L_i$ is reset to 0, and a loop closure link is added between $L_i$ and $L_t$. The loop closure link is used to get neighbors of the old location during Retrieval (Section \ref{sec:retrieval}) and to setup the transition model of the Bayes filter (Section \ref{sec:bayes}). Note that this hypothesis selection differs from our previous work \cite{labbe11memory}: the old parameter $T_{\mathrm{minHyp}}$ is no longer required and locations are not merged anymore on loop closures (only a link is added). Not merging locations helps to keep different signatures of the same location for better hypothesis estimation, which is important in a highly dynamic environment or when the environment changes gradually over time in a cyclic way (e.g., day-night or weather variations).

\subsection{Retrieval}
\label{sec:retrieval}

\begin{figure}[!t] 
\centering 
\includegraphics[width= 3in]{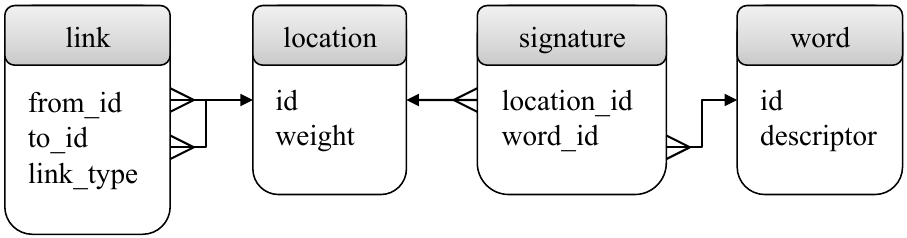} 
\caption[Database schema]{Database representation of the LTM.} 
\label{fig_db_schema} 
\end{figure}

After loop closure detection, neighbors not in WM of the location with the highest loop closure hypothesis are transferred back from LTM to WM. In this work, LTM is implemented as a SQLite3 database, following the schema illustrated in \figurename \ref{fig_db_schema}. In the link table, the \textit{link type} tells if it is a neighbor link or a loop closure link.

When locations are retrieved from LTM, the visual vocabulary is updated with the words associated with the corresponding retrieved signatures. 
Common words from the retrieved signatures still exist in the vocabulary; therefore, a reference is added between these words and the corresponding signatures. 
For words that are no longer present in the vocabulary (because they were removed from the vocabulary when the corresponding locations were transferred [ref. Section \ref{sec:transfer}]), their descriptors are quantized using the same way as in Section \ref{sec:location} (but reusing the kd-trees and doing a linear search on SURF descriptors not yet indexed to kd-trees of the new words added to vocabulary) to check if more recent words represent the same SURF descriptors. 
This step is important because the new words added from the new signature $z_t$ may be identical to the previously transferred words. 
For matched descriptors, the corresponding old words are replaced by the new ones in the retrieved signatures. However, all references in LTM are not immediately changed because this operation is expensive in terms of computational time. Instead, they are changed as other signatures are retrieved, and the old words are permanently removed from LTM when the system is shut down.
If some descriptors are still unmatched, their corresponding words are simply added to vocabulary. 

Algorithm \ref{alg:retrieve} summarizes the Retrieval process.
Because loading locations from the database is time consuming, a maximum of two locations are retrieved at each iteration (chosen inside the neighboring range defined in Section \ref{sec:bayes}). 
When more than two locations can be retrieved, nearby locations in time (direct neighbors of the hypothesis) are prioritized over nearby locations in space (neighbors added through loop closures). 
In \figurename \ref{fig_graph} for instance, if location 116 is the highest loop closure hypothesis, location 118 will be retrieved before location 23. This order is particularly important when the robot is moving, where  retrieving next locations in time is more appropriate than those in space. However, if the robot stays still for some time, all nearby locations in time will be retrieved, followed by nearby locations in space (i.e., 23, 24, 22, 25).  

\begin{algorithm}\small
\begin{algorithmic}[1]
\Procedure{Retrieval}{$L$}
\State $L_r[]\gets$ load a maximum of two neighbors of $L$ from LTM
\Statex\hspace{\algorithmicindent}(with their respective signatures $z_r[]$)
\State Add references to $L_r[]$ for words in $z_r[]$ still in vocabulary 
\State Match old words (not anymore in vocabulary) of $z_r[]$ to
\Statex\hspace{\algorithmicindent}current ones in vocabulary
\State Not matched old words of $z_r[]$ are added to vocabulary
\State Insert $L_r[]$ into WM
\EndProcedure
\end{algorithmic}
\caption{Retrieve neighbors of $L$ from LTM to WM} 
\label{alg:retrieve} 
\end{algorithm}

\subsection{Transfer}
\label{sec:transfer}

When processing time for an image is greater than $T_{\mathrm {time}}$, the oldest locations of the lowest weighted ones are transferred from WM to LTM. 
To be able to evaluate appropriately loop closure hypotheses using the discrete Bayesian filter, neighbors of the highest loop closure hypothesis are not allowed to be transferred. 
The number of these locations is however limited to the finite number of nearby locations in time (accordingly to neighboring range defined in Section \ref{sec:bayes}), to avoid `immunizing' all nearby locations in space (which are indefinite in terms of numbers).
$T_{\mathrm {time}}$ is set empirically to allow the robot to process online the perceived images. 
Higher $T_{\mathrm {time}}$ means that more locations (and implicitly more words) can remain in WM, and more loop closure hypotheses can be kept to better represent the overall environment.
$T_{\mathrm {time}}$ must therefore be determined according to the robot's CPU capabilities, computational load and operating environment. 
If $T_{\mathrm {time}}$ is determined to be higher than the image acquisition time, the algorithm intrinsically uses an image rate corresponding to $T_{\mathrm {time}}$, with 100\% CPU usage.

Because the most expensive step of RTAB-Map is to build the nearest-neighbor index (line 4 of Algorithm \ref{alg:createLocation}), processing time per acquired image can be regulated by changing the vocabulary size, which indirectly influences the WM size.
Algorithm \ref{alg:transfer} presents how the visual vocabulary is modified during the Transfer process. 
A signature of a location transferred to LTM removes its word references from the visual vocabulary. 
If a word does not have reference to a signature in WM anymore, it is transferred into LTM. 
While the number of words transferred from the vocabulary is less than the number of words added from $L_t$ or the retrieved locations, more locations are transferred from WM to LTM. 
At the end of this process, the vocabulary size is smaller than before the new words from $L_t$ and retrieved locations were added, thus reducing the time required to create the nearest-neighbor index (kd-trees) from the vocabulary for the next image. 
Saving the transferred locations into the database is done asynchronously using a background thread, leading to a minimal time overhead for the next iteration when joining the thread on line 17 of Algorithm 1 ($pTime$ then includes transferring time).

The way to transfer locations into LTM influences long-term operation when WM reaches its maximum size, in particular when a region (a set of locations) is seen more often than others. 
Normally, at least one location in a new region needs to receive a high weight through the Weight Update to replace an old and high weighted one in WM, in order to detect loop closures when revisiting this region. However, if there is no location in the new region that receives a high weight, loop closures could be impossible to detect unless the robot comes back to a high weighted location in an old region, and then moves from there to the new one. To handle this situation and to improve from our previous work \cite{labbe11memory}, a subset of the highest weighted locations (defined by $T_{\mathrm{recent}}\times N_{\mathrm {WM}}$) after the last loop closure detected are not allowed to be transferred to LTM. 
This way, when the robot explores a new region, there are always high weighted locations of this region in WM until a loop closure is detected.
If the number of locations after the last loop closure detected in WM exceeds $T_{\mathrm {recent}}\times N_{\mathrm {WM}}$, these locations can be transferred like the ones in WM (i.e., with the criterion of the oldest of the lowest weighted locations).
$T_{\mathrm {recent}}$ is a ratio fixed between 0 and 1. A high $T_{\mathrm {recent}}$ means that more locations after the last loop closure detected are kept in WM, which also leads to a transfer of a higher number of old locations to LTM. 

\begin{algorithm}\small
\begin{algorithmic}[1]
\Procedure{Transfer}{ }
\State $nwt\gets 0$ \Comment number of words transferred
\State $nwa \gets$ number of new words added by $L_t$ and retrieved
\Statex\hspace{\algorithmicindent}locations
\While{ $nwt<nwa$}
\State $L_i\gets$ select a transferable location in WM (by weight
\Statex\hspace{\algorithmicindent}\hspace{\algorithmicindent}and age), ignoring retrieved locations and those in recent
\Statex\hspace{\algorithmicindent}\hspace{\algorithmicindent}WM (using $T_{\mathrm{recent}}$)
\State Move $L_i$ to $trash$
\State Move words $w_i$ which have no more references to any
\Statex\hspace{\algorithmicindent}\hspace{\algorithmicindent}locations in WM to $trash$
\State $nwt\gets nwt +\Call{size}{w_i}$
\EndWhile
\State Start $trash$'s thread to empty $trash$ to LTM
\EndProcedure
\end{algorithmic}
\caption{Transfer locations from WM to LTM}  
\label{alg:transfer} 
\end{algorithm}

\section{Results}
\label{sec:results}
\label{results}

Performance of RTAB-Map is evaluated in terms of precision-recall metrics \cite{cummins2009highly}. Precision is the ratio of true positive loop closure detections to the total number of detections. Recall is defined as the ratio of true positive loop closure detections to the number of ground truth loop closures. 
To situate what can be considered good recall performance, for metric SLAM, recall of around 20\% to 30\% at 100\% precision (i.e., with no false positives) is sufficient to detect most loop closure events when the detections have uniform spatial distributions \cite{cummins2009highly}. 
Note however that the need to maximize recall depends highly on the SLAM method associated with the loop closure detection approach. If metric SLAM with excellent odometry is used, a recall ratio of about 1\% could be sufficient.
For less accurate odometry (and even no odometry), a higher recall ratio would be required. 

Using a MacBook Pro 2.66 GHz Intel Core i7 and a 128 Gb solid state hard drive, experimentation is done on ten community data sets and two custom data sets using parameters presented in Table \ref{table_parameters}. These parameters were set empirically over all data sets to give good overall recall performances (at precision of 100\%), and remained the same (if not otherwise stated) to evaluate the adaptability of RTAB-Map.
The only SURF parameter changed between experiments is $T_{\mathrm{response}}$, which is set based on the image size. $T_{\mathrm{time}}$ is set accordingly to image rate of the data sets. As a rule of thumb, $T_{\mathrm {time}}$ can be about 200 to 400 ms smaller than the image acquisition rate at 1 Hz, to ensure that all images are processed under the image acquisition rate, and even if the processing time goes over $T_{\mathrm {time}}$ ($T_{\mathrm {time}}$ then corresponds to the average processing time of an image by RTAB-Map). So, for an image acquisition rate of 1 Hz, $T_{\mathrm {time}}$ can be set between 600 ms to 800 ms.
For each experiment, we identify the minimum $T_{\mathrm{loop}}$ that maximizes recall performance at 100\% precision. With the use of these bounded data sets, the LTM database could have been placed directly in the computer RAM, but it was located on the hard drive to simulate a more realistic setup for timing performances. Processing time $pTime$ is evaluated without the SURF features extraction step (lines 2-3 of Algorithm \ref{alg:createLocation}), which is done by the camera's thread in the implementation.

\begin{table}[!t]
\renewcommand{\arraystretch}{1.3}
\caption{RTAB-Map and SURF Parameters}
\label{table_parameters}
\centering
\begin{tabular}{ll|ll}
\hline
$T_{\mathrm{STM}}$ & 30 &  SURF dimension & 64\\
$T_{\mathrm{similarity}}$ & 20\% & SURF $T_{\mathrm{NNDR}}$ & 0.8\\
$T_{\mathrm{recent}}$ & 20\% & SURF $T_{\mathrm{maxFeatures}}$ & 400\\
&  & SURF $T_{\mathrm{bad}}$ & 0.25\\
\hline
\end{tabular}
\end{table}

\subsection{Community Data Sets}
\label{community}

We conducted tests with the following community data sets: NewCollege (NC) and CityCentre (CiC) \cite{Cummins08a}; Lip6Indoor (L6I) and Lip6Outdoor (L6O) \cite{Angeli08c}; 70 km \cite{cummins2009highly}; NewCollegeOmni (NCO) \cite{smith2009new}; CrowdedCanteen (CrC) \cite{Kawewong2011pirfnav2}; BicoccaIndoor-2009-02-25b (BI), BovisaOutdoor-2008-10-04 (BO) and BovisaMixed-2008-10-06 (BM) \cite{ceriani2009rawseeds} data sets. NC and CiC data sets contain images acquired from two cameras (left and right), totaling 2146 and 2474 images respectively of size $640 \times 480$. Because RTAB-Map takes only one image as input in its current implementation, the images from the two cameras were merged into one, resulting in 1073 and 1237 images respectively of size $1280 \times 480$.
For data sets with an image rate depending on vehicle movement or over 2 Hz, some images were removed to have approximately an image rate of 1 Hz (i.e., keeping 5511 of the 9575 panoramic images for the 70 km data set). 
For NCO and CrD data sets, because they contain panoramic images taken by a vehicle slower than for the 70 km data set, $T_{\mathrm{similarity}}$ is increased to 35\% (compared to 20\% for all other data sets). $T_{\mathrm{time}}$ is set  to $1.4 \, \mathrm{s}$ for images acquired every 2 seconds (0.5 Hz), to $0.7 \, \mathrm{s}$ for images acquired every second (1 Hz) and to $0.35 \, \mathrm{s}$ for images acquired every half second (2 Hz).

\begin{table*}[!t]
\renewcommand{\arraystretch}{1.3}
\caption{Experimental conditions and results of RTAB-Map on community data sets}
\label{table_community_results}
\centering
\begin{tabular}{l|c|c|c|c|c|c|c|c|c|c}
\hline
Data set & 
NC &
CiC & 
L6I & 
L6O & 
70 km &
NCO &
CrC & 
BI & 
BO & 
BM \\

\# images & 
1073 & 
1237  & 
388 &
531  &
5511  &
1626  & 
692  & 
1757  &
2277 &
2147 \\

Image size & 
1280x480 & 
1280x480 & 
240x192 &
240x192 &
1600x460\footnotemark[2] &
2048x618 & 
480x270 & 
320x240 &
320x240 &
320x240\\

Image rate & 
$\approx$0.5 Hz & 
$\approx$0.5 Hz & 
1 Hz &
0.5 Hz &
$\approx$1 Hz &
1 Hz & 
2 Hz & 
1 Hz &
1 Hz &
1 Hz\\
\hline

$T_{\mathrm{response}}$ & 
1000 & 
1000 & 
10 & 
10 &
1000 &
1000 & 
75 & 
50 & 
50 &
50 \\

$T_{\mathrm{time}}$ (s) & 
1.4 & 
1.4 & 
0.7 & 
1.4 &
0.7 &
0.7 & 
0.35 & 
0.7 & 
0.7 &
0.7
\\
\hline

Max $pTime$ (s) & 
1.77 & 
1.73 & 
0.74 &
1.58 &
0.94 & 
0.94 & 
0.44 & 
0.86 &
0.87 &
0.87\\

Max dict. size $\times10^3$ & 
110 &
112 &
52 &
111 &
57 & 
56 &
28 &
57 &
57 &
57\\

Max WM size & 
410 & 
377 & 
334 &
373 &
162 &
259 & 
101 & 
220 &
210 &
204 \\

Min $T_{\mathrm{loop}}$ & 
0.11 & 
0.08 & 
0.14 &
0.07 &
0.11 &
0.10 & 
0.09 & 
0.43 &
0.17 &
0.18 \\

Precision (\%) & 
100 & 
100 & 
100 &
100 &
100 &
100 & 
100 & 
100 &
100 &
100\\

Recall (\%) & 
\textbf{89} & 
\textbf{81} & 
\textbf{98} &
\textbf{95} &
\textbf{59} &
\textbf{92} & 
\textbf{95} & 
\textbf{82} &
\textbf{56} &
\textbf{72} \\
\hline

Recall (\%) from other & 
\multirow{2}{*}{47 \cite{Cummins08a}} & 
37 \cite{Cummins08a} & 
78 \cite{Kawewong2011pirfnav2} &
\multirow{2}{*}{71 \cite{Angeli08c}} &
\multirow{2}{*}{49 \cite{cummins2009highly}} &
$\approx$7 \cite{newman2009navigating} & 
\multirow{2}{*}{87 \cite{Kawewong2011pirfnav2}} & 
\multirow{2}{*}{58 \cite{galvez2011real}} &
\multirow{2}{*}{6 \cite{galvez2011real}} &
\multirow{2}{*}{28 \cite{galvez2011real}} \\
approaches [ref. \#]  & & 80 \cite{Kawewong2011pirfnav2} & 80 \cite{Angeli08c} & & & 38 \cite{maddern2011continuous} & & & & \\
\hline
\end{tabular}
\end{table*}

\begin{figure}[!t] 
\centering 
\includegraphics[width= 2.2in]{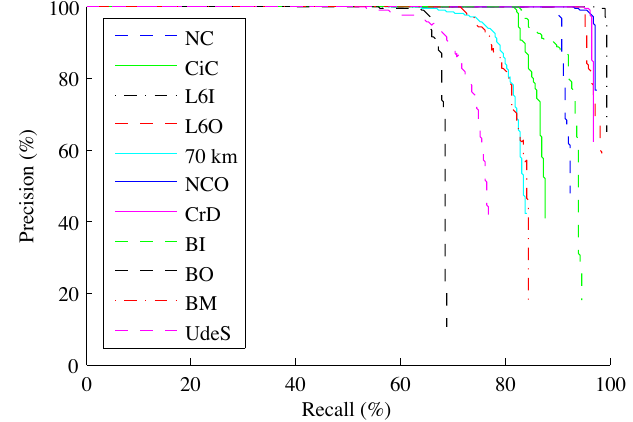} 
\caption[Precision-recall curves]{Precision-recall curves for each data set.} 
\label{fig:Precision_Recall} 
\end{figure}

Table \ref{table_community_results} summarizes experimental conditions and results. Recall performance corresponds to the maximum recall performance observed at 100\% precision, and precision-recall curves are shown in \figurename \ref{fig:Precision_Recall}. Compared to other approaches that also used these data sets, RTAB-Map achieves better recall performances at 100\% precision (with improvements up to 54\%, as shown at the bottom of Table \ref{table_community_results}) while respecting real-time constraints: the maximum processing time $pTime$ is always under the image acquisition time.

\subsection{Universit\'e de Sherbrooke (UdeS) Data Set}
\label{sec:udes}

\begin{figure}[!t] 
\centering 
\includegraphics[width= 2.77in]{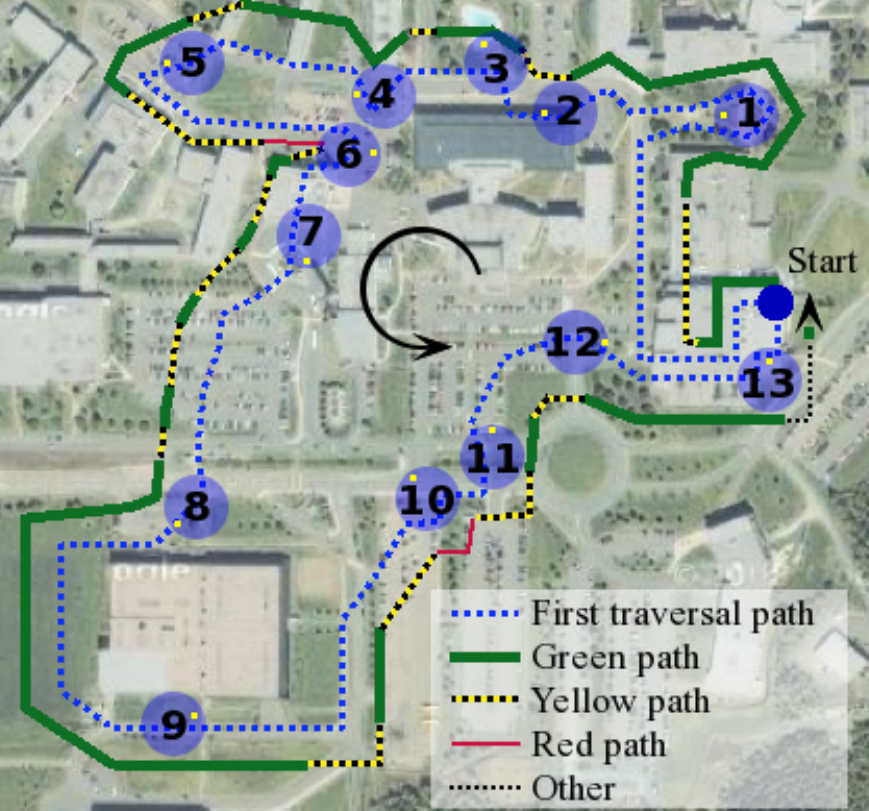} 
\caption[UdeS data set aerial view]{UdeS data set aerial view. The first traversal is represented by the dotted line. The second traversal is represented by a line located around the first. The start/end point is represented by the circle. The small white dots in the waypoint ID numbers represent camera orientation at this location. Recall performance is from the test case with $T_{\mathrm{time}}=0.7 \, \mathrm{s}$.} 
\label{fig_udes_data_set_results} 
\end{figure}

The data set used for this test is made of images taken over a 2 km loop of the Universit\'e de Sherbrooke (UdeS) campus, traversed twice, as illustrated by \figurename \ref{fig_udes_data_set_results}. A total of 5395 images of $640 \times 480$ resolution at 1 Hz were captured with a handheld webcam, over 90 minutes. The data set contains a variety of environment conditions: indoor and outdoor, roads, parkings, pedestrian paths, trees, a football field, with differences in illumination and camera orientation. To study the ability of RTAB-Map to transfer and to retrieve locations based on their occurrences, we stopped at 13 waypoints during which the camera remained still for around 20 to 90 seconds, leading to 20 to 90 images of the same location. After processing the images of the first traversal, it is expected that the related locations will still be in WM and that RTAB-Map will be able to retrieve nearby locations from LTM to identify loop closures. $T_{\mathrm{response}}$ is set to $150$.

\begin{table*}[!t]
\renewcommand{\arraystretch}{1.3}
\caption{Experimental results of RTAB-Map for the UdeS data set (5395 images of size 640x480 taken at 1 Hz)}
\label{table_udes_results}
\centering
\begin{tabular}{l|c|c|c|c|c|c|c|c|c|c|c|c|c|c|c|c}
\hline
$T_{\mathrm{time}}$ (s) & 
$\infty$ & 0.95 & 0.90 & 0.85 & 0.80 & 0.75 & \textbf{0.70} & 0.65 & 0.60 & 0.55 & 0.50 & 0.45 & 0.40 & 0.35 & 0.30 & 0.25\\
\hline
Max $pTime$ (s) & 
 10.9 & 1.11 & 1.08 & 1.05 & 1.01 & 0.94 & \textbf {0.87} & 0.81 & 0.74 & 0.72 & 0.61 & 0.56 & 0.50 & 0.46 & 0.37 & 0.32 \\
Max dict. size $\times10^3$ & 
714 & 76 & 73 & 68 & 65 & 60 & \textbf {57} & 52 & 49 & 44 & 41 & 36 & 33 & 29 & 24 & 21 \\
Max WM size & 
2870 & 302 & 280 & 265 & 256 & 240 & \textbf{221} & 205 & 184 & 169 & 155 & 136 & 118 & 103 & 85 & 69 \\
Min $T_{\mathrm{loop}}$ &  
0.19 & 0.11 & 0.09 & 0.09 & 0.09 & 0.10 & \textbf {0.10} & 0.11 & 0.11 & 0.10 & 0.15 & 0.12 & 0.12 & 0.12 & 0.10 & 0.14 \\
Precision (\%) &  
100 & 100 & 100 & 100  & 100 & 100 & \textbf {100} & 100 & 100 & 100 & 100 & 100 & 100 & 100 & 100 & 100 \\
Recall (\%) &
 34 & 47 & 48 & 51 & 54 & 48 & \textbf {51} & 49 & 50 & 53 & 41 & 44 & 47 & 39 & 34 & 11\\
\hline
\end{tabular}
\end{table*}

Table \ref{table_udes_results} presents results using different $T_{\mathrm{time}}$ to show the effect of memory management on recall performances. When $T_{\mathrm {time}}=\infty$, all locations are kept in WM; therefore, loop closure detection is done over all previously visited locations. The resulting maximum processing time is 10.9 seconds, which makes it impossible to use online (image acquisition time is 1 sec).
With $T_{\mathrm{time}}<=0.75$ s, processing is done online (i.e., processing time is always lower than the image acquisition time). 

For $T_{\mathrm{time}}\in[0.95;0.35] \, \mathrm{s}$, recall varies between 39\% and 54\%, and this variation is caused by the choice of locations kept in WM: small computation time variations and different $T_{\mathrm{time}}$ explain why some locations are transferred or retrieved in some experiments while they are not in others. Lower recall performance at $T_{\mathrm{time}}=\infty$ is caused by the presence of a large vocabulary: as words are added to vocabulary, the matching distance between SURF features becomes smaller because of the $T_{\mathrm{NNDR}}$ matching criterion. Although this provides more precise matches, the matching process is more sensitive to noise. 
Note also that with a larger WM, there are more chances that a new location (in presence of dynamic environmental changes) would also be found similar to another old location still in WM (which would have been in LTM with a smaller WM), creating more false positives.
Conducting tests to determine the optimal size of WM for the best recall could be done, but this would be highly dependent on the environment and the path followed by the robot, and would not necessarily satisfy real-time constraints.
For $T_{\mathrm{time}}<0.35 \, \mathrm{s}$, the WM size becomes too small, and RTAB-Map is unable to detect as many loop closures: when $T_{\mathrm{time}}$ is reached (because the retrieved locations cannot be immediately transferred), old locations with large weights are transferred instead, making it difficult to detect loop closures if Retrieval does not bring back appropriate locations.

\begin{figure}[!t] 
\centering 
\includegraphics[width= 3.1in]{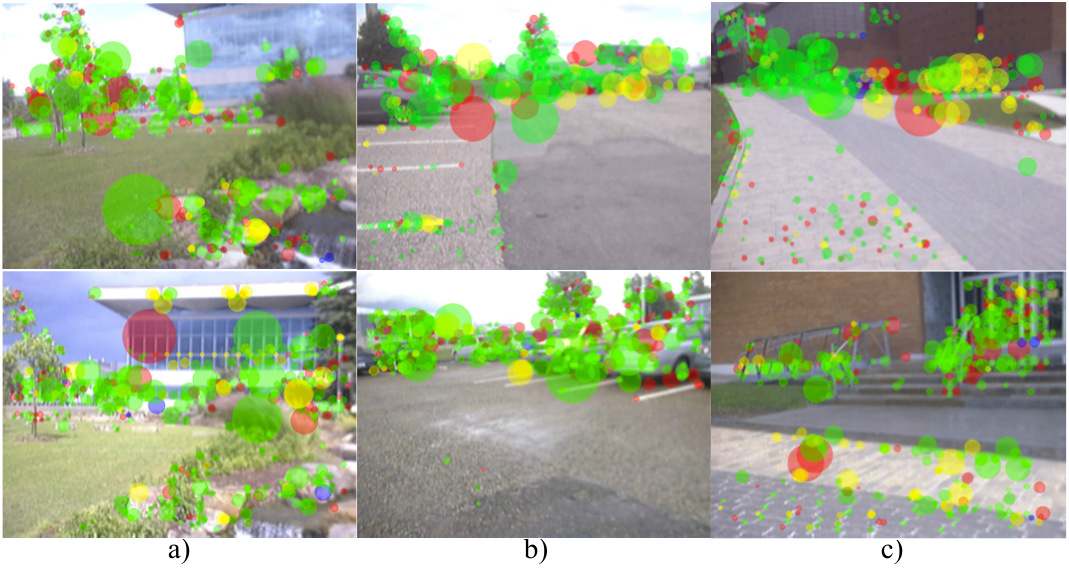} 
\caption[Examples of rejected hypotheses]{Examples (with visual words represented as circles) of loop closure hypotheses that are insufficient (under $T_{\mathrm{loop}}$) to detect loop closures, caused by changes in illumination conditions (a) or camera orientation (b and c). New words are colored in green or yellow, and words already in the vocabulary are colored in red or blue.} 
\label{fig_failed_samples} 
\end{figure}

\begin{figure}[!t] 
\centering 
\includegraphics[width= 2.4in]{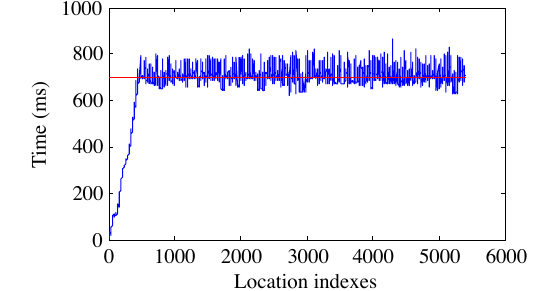}  
\caption[Processing time]{Processing time in terms of locations processed over time with the UdeS data set. $T_{\mathrm{time}}$ is set at $0.7 \, \mathrm{s}$, as shown by the horizontal line.} 
\label{fig_time_result} 
\end{figure}

\footnotetext[2]{The 5-view omni-directional images were stitched using The Panorama Factory software.}
%

To illustrate more closely the results obtained for $T_{\mathrm{time}}=0.7 \, \mathrm{s}$, at the end of the trial each waypoint is represented in WM by at least one location with a high weight. The other locations with smaller weights that are still in WM are the ones where the images did not change much over time (like the football field). At its maximum, the vocabulary has 57K visual words for 251 locations (221 in WM + 30 in STM). With $T_{\mathrm{maxFeatures}}=400$, there are then about 227 unique words and 173 common words per location. The high number of unique words is attributable to $T_{\mathrm{NNDR}}$ criterion on feature quantization. 
\figurename \ref{fig_udes_data_set_results} illustrates recall performance over the 2 km loop, categorized using three colored paths:

\begin{itemize}
\item \textbf{Green} paths identify valid loop closure hypotheses. Ground truth was labeled manually, based on similar visual appearance and proximity between images.
The images from the second traversal are not taken at exactly the same position and the same angle compared to the first traversal. Therefore, a match within a margin of 10 locations is considered acceptable for loop closure.

\item \textbf{Yellow} paths indicate an insufficient loop closure probability under $T_{\mathrm{loop}}$. 
However, a Yellow path also means that Retrieval works as expected, i.e., RTAB-Map is able to retrieve appropriate transferred signatures (those near ground truth loop closures) as previously visited locations are encountered. \figurename \ref{fig_failed_samples} a) illustrates an example: a change in illumination conditions from the first traversal (top image) caused a change in the visual words extracted from the image of the second traversal (bottom image). Most of the words extracted in the top image are from the tree on the left, compared to the building section in the bottom image.

\item \textbf{Red} paths identify when there is no location in WM which could be matched with the current location. 
A transition from a Yellow to a Red path occurs as follows. The likelihood of the observed location with the corresponding location of the first traversal still in WM is too low (i.e., words extracted are too different or are too common) and is lower than with other locations in the map. Because the associated loop closure hypothesis is not the highest one anymore, nearby locations of the real loop closure cannot be retrieved from LTM to WM. Therefore, the next observed locations do not have any locations in WM which can be used to find loop closures. \figurename \ref{fig_failed_samples} b) and c) illustrate what happens at the beginning of two Red paths (before the waypoints 6 and 10 respectively): over several consecutive images, the camera was not oriented in the same direction as in the first traversal, and RTAB-Map was not able to retrieve neighbor locations from LTM because the new acquired locations were more similar to locations in another part of the map. However, the wrong loop closure hypotheses (or false positives) during a Red path stayed low under $T_{\mathrm{loop}}$, and thus they were not accepted.
A transition from a Red to a Yellow path happens when the camera returns to a location still in WM, increasing the associated loop closure hypothesis to become the highest and resulting in a retrieval of neighbor locations from which loop closures can be detected.

\item \textbf{Other} indicates paths different from the ones taken during the first traversal, and therefore there are no loop closures to find.

\end{itemize}
Finally, \figurename \ref{fig_time_result} shows the processing time for each acquired image with $T_{\mathrm{time}}=0.7 \, \mathrm{s}$. As expected, once the processing time has reached $T_{\mathrm{time}}=0.7 \, \mathrm{s}$ (after 444 images), the memory management is triggered and the processing time remains close to $T_{\mathrm{time}}$, with an average processing time of $0.67 \, \mathrm{s}$ and a maximum processing time of $0.87 \, \mathrm{s}$.

\subsection{``Need for Speed: Most Wanted'' (NFSMW) Data Set}
\label{sec:enclosedExperiment}

\begin{figure}[!t] 
\centering 
\includegraphics[width= 1.7in]{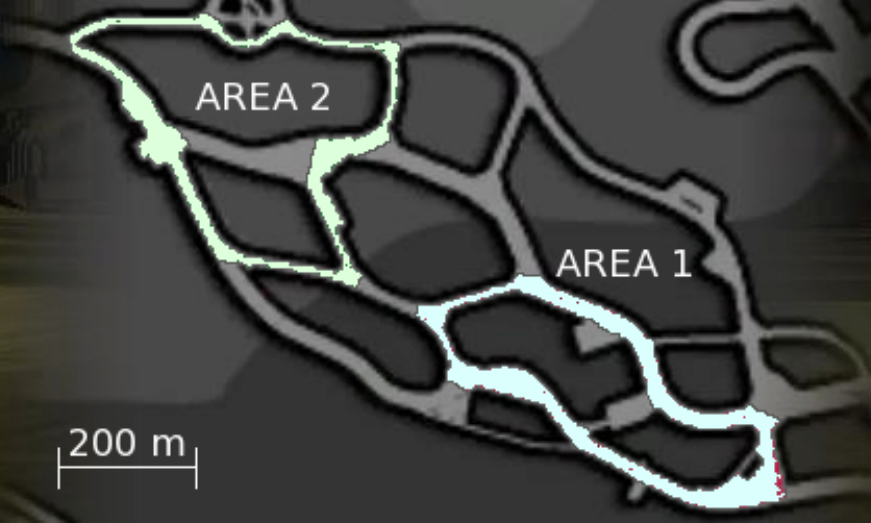} 
\caption[Map]{Map of NFSMW data set showing Area 1 and Area 2.} 
\label{fig:map_nfs} 
\end{figure}

\begin{figure}[!t] 
\centering 
\includegraphics[width= 3.2in]{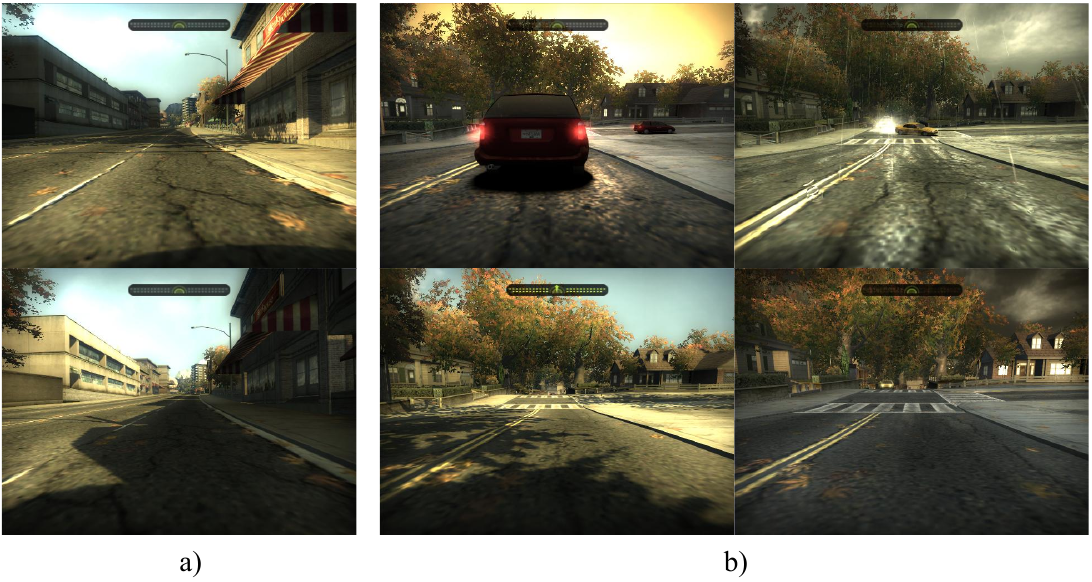} 
\caption[NFS samples]{Samples of the NFSMW data set. In a), the sun shines come from the east (bottom) or the west (top); note the high color and contrast differences for the left and right buildings caused by the dynamic shadows. In b), four different atmospheric conditions are shown for the same location (over 30 minutes).} 
\label{fig:nfs_samples} 
\end{figure}

For this experiment, the video game ``Need for Speed : Most Wanted'' (NFSMW) was used to acquire images by driving around (while respecting speed limits) city Area 1 about one hundred times and then moving to Area 2 for another hundred traversals. \figurename \ref{fig:map_nfs} illustrates these two areas. This data set was generated to evaluate RTAB-Map in two specific conditions:

\begin{enumerate}
\item Frequently observing the same locations;

\item Moving to new locations after observing the same area for a long period of time.
\end{enumerate}
Even though the environment is synthetic, there are many large changes in illumination conditions (sun and shadows move slowly; there are also bright sunrises and random storms) that makes it very challenging for loop closure detection over long-term operation. \figurename \ref{fig:nfs_samples} illustrates examples of such changes. A total of 25098 images of 640$\times$480 size were taken at 1 Hz, totalizing 7 hours of driving. Because of the presence of the head-up display in the images and that the lower portions are generally made of common and repetitive road textures, the upper 10\% and lower 40\% of the images are not used for SURF features extraction. $T_{\mathrm{response}}$ is set to $150$. 

\begin{figure}[!t] 
\centering 
\includegraphics[width= 3.2in]{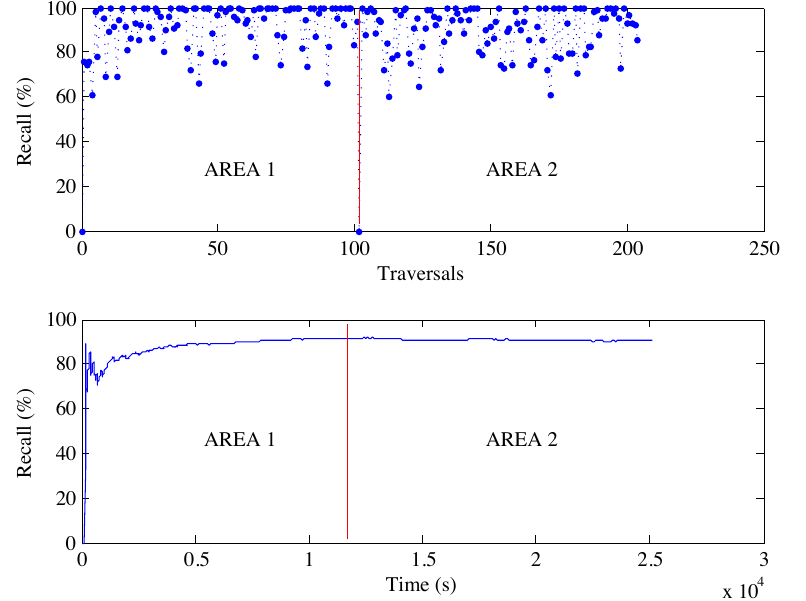} 
\caption[Recall over time]{Recall performance over traversals (top) and time from $t=0$ (bottom) for the NFSMW data set (at 100\% precision).} 
\label{fig:RecallOverTime} 
\end{figure}

The upper portion of \figurename \ref{fig:RecallOverTime} illustrates recall performances computed for each traversal. The performances for the 102 traversals of Area 1 and the 103 traversals of Area 2 are delimited by the red vertical line. Recall of 0\% is observed for the first traversal of Area 1 and the first traversal of Area 2, as expected. Recall variations are caused by changes in illumination conditions during the traversals : if a storm happens when sunny locations are retrieved, loop closures are not detected until the storm finishes or a darker version of the encountered locations are retrieved. Generally, Red paths end on road intersections, which corresponds to locations with higher weights (caused by having the vehicle stop) and where neighbors (by loop closures) with different illumination conditions are retrieved. 
For Area 1 and Area 2, recall at 100\% precision varies from 60\% to 100\%.

The lower portion of \figurename \ref{fig:RecallOverTime} shows recall performances over time at 100\% precision, considering all loop closures detected from $t=0$. 
After encountered most of the changes in illumination conditions between $t=0$ to $t=5000$ (which corresponds to 41 traversals), the average recall performance stabilizes around 89\%, for a minimum $T_{loop}=0.10$. With a ground truth of 24800 loop closures and 89\% recall, there are 2728 duplicated locations in the global graph of the 280 locations of Area 1 and Area 2. These duplicated locations create new paths in the global map. \figurename \ref{fig_graph} illustrates such a case: locations 453 and 454 are duplicates of 114 and 115 respectively because the environment changed too much; location 455 then have two paths representing the same real location. In practice, it is likely that one of the paths will eventually be transferred in LTM, keeping only one version of the real location in WM. However, keeping two paths representing the same locations in WM may be beneficial, especially in dynamical environments with cyclic atmospheric changes like in this data set: locations in dark conditions could have almost no features similar to their versions in bright conditions, then loop closures are found alternately between dark and bright versions. 

Looking more closely at the transition between the two areas, when moving to Area 2, the WM only contains locations of Area 1 with large weights. If a set of the new locations would not have been kept in the recent part of WM (as explained in Section \ref{sec:transfer}), loop closures would have been impossible to detect if no location received a high weight (from Weight Update) to replace locations in WM from Area 1. After the first traversal of the Area 2, the recent part of WM was populated mostly by locations representing road intersections (having higher weights because the vehicle stopped). The first loop closure detected on the second traversal was found on the first intersection encountered during the first traversal of Area 2. Next locations were then retrieved, and a recall of 100\% at 100\% precision was achieved for the second traversal. 

Regarding processing time, once $T_{\mathrm{time}} = 0.7 \, \mathrm{s}$ is reached, a mean time of $0.71 \, \mathrm{s}$ is achieved for the rest of the experiment. The maximum processing time is $0.93 \, \mathrm{s}$, thus respecting the real-time constraint of $1$ Hz. At the end of the experiment, there were 19 locations of the Area 1 and 38 locations of the Area 2 with high weights in WM, distributed mainly on road intersections.

\section{Discussion}
\label{sec:discussion}
Results presented in Section \ref{sec:results} suggest that RTAB-Map can achieve good recall performances at 100\% precision over diverse and large-scale environments. 
Real-time constraints can be satisfied independently of the scale of the environment, which is very important for long-term online mapping. 

Overall, using similarity occurrences reveal to be a simple and functional method to determine which locations to keep in WM. Obviously, it has limitations when an area is seen only one time before moving to a new area for a long time. 
To illustrate, we conducted a trial using the NFSMW setup by doing only one traversal of Area 1 and then moving for one hundred traversals of Area 2. After the first traversal of Area 1, the highest weight of a location is 2. After 56 traversals of Area 2, all locations of Area 1 were transferred to LTM (in comparison to 19 locations remaining in WM after 103 traversals of Area 2 in Section \ref{sec:enclosedExperiment}). The number of traversals of Area 2 required to transfer all locations of Area 1 in LTM depends on the weight assigned to locations during the first traversal.
Returning to Area 1 then leads to the creation of duplicated locations that cannot be associated to the locations of Area 1 stored in LTM (unless they were revisited backward from the entry point of Area 2), and these duplicated locations, if visited frequently, can remain in WM to be used in future loop closures.
This illustrates the compromise to be made to satisfy real-time constraints: it may happen that infrequently visited locations get transferred to LTM without being able to be remembered back, but at the same time such locations are not used in the loop closure detection process, allowing to speed up the process using only locations that have more chance to be revisited. Such compromise is therefore driven by the environment and the experiences of the robot. Note that other methods to assign weights to locations could be imagined, such as having the system identify which locations are important (and assigning directly a high weight to these locations) based on events, the robot's internal states or even from user inputs. Also, approaches such as sparsely or randomly sampling the LTM could be used to prevent forgetting entire areas  not visited often enough.

In RTAB-Map, LTM's growth influences loop closure detection performance over large-scale and long-term operation. To understand such influence, let's define $w_w$, the number of words in WM and STM (i.e., the visual vocabulary), $n_w$, the number of locations in WM, $w_l$, the number of words in LTM and $n_l$, the number of locations in LTM. Time complexity for each step of Algorithm \ref{alg:mainLoop} is given as follows:

\begin{itemize}
\item Location Creation: building the kd-trees from the vocabulary is $O(w_w\mathrm{log}w_w)$, and quantizing SURF descriptors extracted from the new image to kd-trees is $O(\mathrm{log}w_w)$. The SURF features extraction can be considered $O(1)$ as image size is fixed.

\item Weight Update: updating weight is $O(1)$.

\item Bayesian Filter Update: computing observation is $O(n_w)$ and belief if $O(n_w^2)$.

\item Loop Closure Hypothesis Selection: hypothesis selection is $O(n_w)$.

\item Retrieval: SURF descriptors quantization is $O(\mathrm{log}w_w)$. Database selection query is $O(\mathrm{log}[w_l+n_l])$.

\item Transfer: selecting a transferrable location is $O(n_w)$. Database insertion query is $O(\mathrm{log}[w_l+n_l])$.

\end{itemize}

\begin{figure}[!t] 
\centering 
\includegraphics[width= 2.6in]{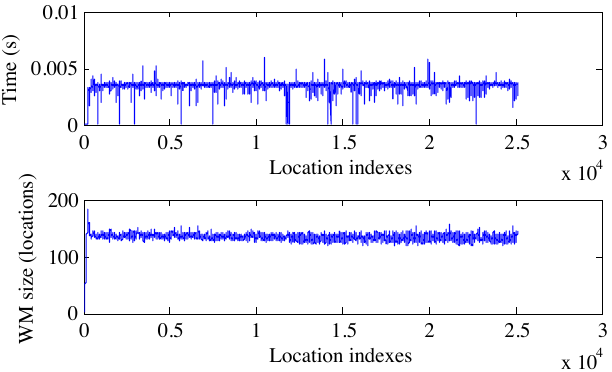} 
\caption[Database time and WM size]{(Top) Total database (LTM) access time to retrieve and transfer locations per iteration. (Bottom) WM size variation during the experiment.} 
\label{fig_db_wm} 
\end{figure}

When $T_{\mathrm{time}}$ is reached, WM size remains fixed, bounding time complexities associated to $w_w$ and $n_w$. However, for Retrieval and Transfer, time complexities also depend on LTM, and LTM size is not bounded. With the growth of LTM, $T_{\mathrm{time}}$ is more likely to be reached, and WM size will gradually decrease over time to satisfy real-time constraints. Theoretically, WM size may eventually become null, disabling loop closure detection. In practice, though, the logarithmic growth in time complexity caused by LTM is very small and WM size is not affected.
For the NFSMW experiment, top of \figurename \ref{fig_db_wm} shows the total database access time required to retrieve and to transfer locations for each RTAB-Map iteration, for up to 7 hours of use. Time growth is unnoticeable. At the end of the experiment, the database size is $3.1$ GB with 6.3 million words and 25098 locations (all merged and bad locations were kept in the database for debugging purpose). The bottom of \figurename \ref{fig_db_wm} shows the WM size over time. The higher variations of WM size after around 12000 locations are mainly caused by environmental changes from Area 1 to Area 2. 
If necessary, a solution to LTM size would be to limit database growth by permanently removing offline some locations from the database. For instance, paths leading to the same high weighted locations could be eliminated based on the sum of the weights of locations in the paths.
If the number of distant high weighted locations gets very high, important locations could ultimately be deleted, resulting in a dismembered global map (i.e., many disconnected smaller maps) if weight is the primary transfer criterion. Pruning the oldest locations (independently of the weight) may be better to preserve a unique global map, at the cost of forgetting important old locations.

In dynamic environments, the performance of RTAB-Map is also highly dependent on the quality of the SURF features extracted. We observed in our experiments that SURF features are relatively sensible to changes in illumination and shadows, reducing the number of discriminative features for more ``garbage features'' (or common words) in images. In RTAB-Map, at least one discriminative feature in the environment is required to find a loop closure, but if there are many ``garbage features'', this means that other locations also receive high likelihoods, thus shadowing the weight of the discriminative feature. In this case, Eq. \ref{eq:virtualPlaceLikelihood} scores high (for a new location probability) because the standard deviation of the likelihood scores is small comparatively to the mean.
Feature weighting may help in such cases by assigning a high weight to discriminative features and a lower weight to ``garbage features''. However, we think that doing so would lead to more false positives. By considering all features with the same weight in RTAB-Map, many discriminative features are required for a location to score higher than others if many ``garbage features'' are present, decreasing the chance of false positives. We prefer avoiding to find a loop closure in such condition (like in environments populated with many dynamic objects or people) rather than increasing the probability to accept a false positive.

As additional improvements, exploiting sparseness of the Bayesian filter \cite{cummins2009highly} or using more efficient nearest-neighbor structures (to avoid reconstructing the whole kd-trees at each iteration) may speed up the process to keep more locations in WM. However, our focus in this paper is on real-time constraints satisfaction (i.e., what should be done when computation time reaches the time threshold), and not optimizing complexities depending on WM size.  
Finally, to overcome the occurrences of Red paths caused by changes in camera orientation (see Section \ref{sec:udes}), active localization could be triggered by detecting decreasing hypotheses, which could make the system move the camera in the right direction to let RTAB-Map retrieve appropriate locations from LTM to WM.

\section{Conclusion}
\label{sec:conclusion}
Results presented in this paper suggest that RTAB-Map, a loop closure detection approach based on a memory management mechanism, is able to meet real-time constraints needed for online large-scale and long-term operation. 
While keeping a relatively constant number of locations in WM, online processing is achieved for each new image acquired. 
Retrieval is a key feature that allows RTAB-Map to reach adequate recall ratio even when transferring a high proportion of the perceived locations in LTM, which are not used for loop closure detection.
In future work, in addition to possible extensions outlined in Section \ref{sec:discussion}, we plan to study how RTAB-Map can be combined to other approaches to implement a complete Simultaneous Localization and Mapping system.

\bibliographystyle{IEEEtran}
\bibliography{IEEEabrv,../../../Papers/References}

\begin{IEEEbiography}[{\includegraphics[width=1in,height=1.25in,clip,keepaspectratio]{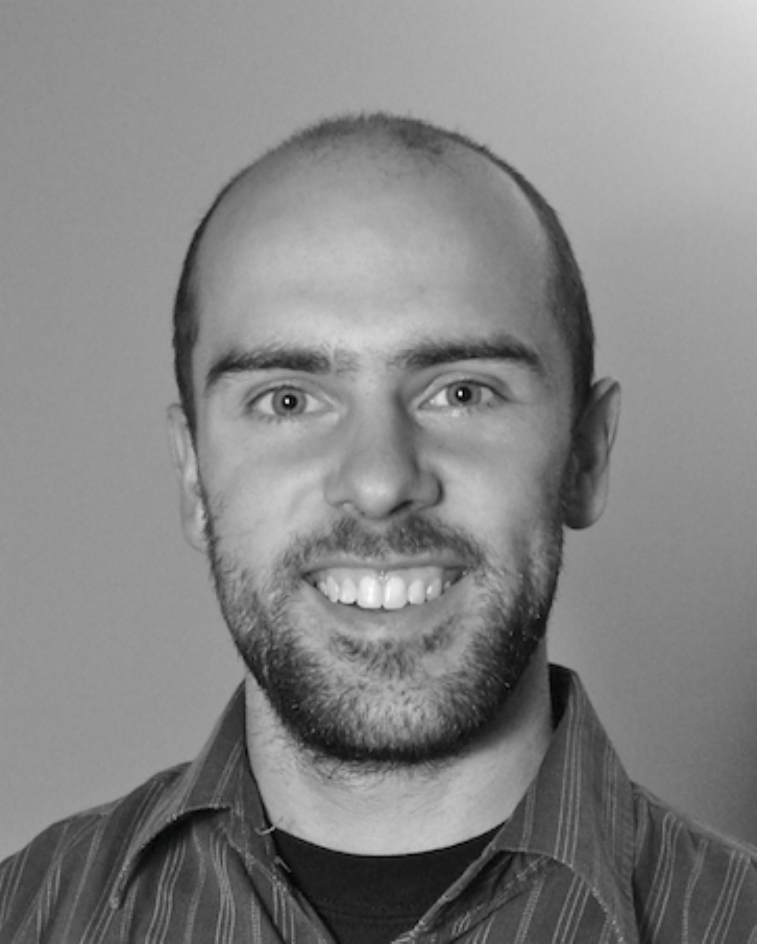}}]
{Mathieu Labb\'e}
received the B.Sc.A. degree in computer engineering and the M.Sc.A. degree in electrical engineering from the Universit\'e de Sherbrooke, Sherbrooke, Qu\'ebec Canada, in 2008 and 2010, respectively. He is currently working toward the Ph.D. degree in electrical engineering at the same university.

His research interests include computer vision, autonomous robotics and robot learning.
\end{IEEEbiography}

\begin{IEEEbiography}[{\includegraphics[width=1in,height=1.25in,clip,keepaspectratio]{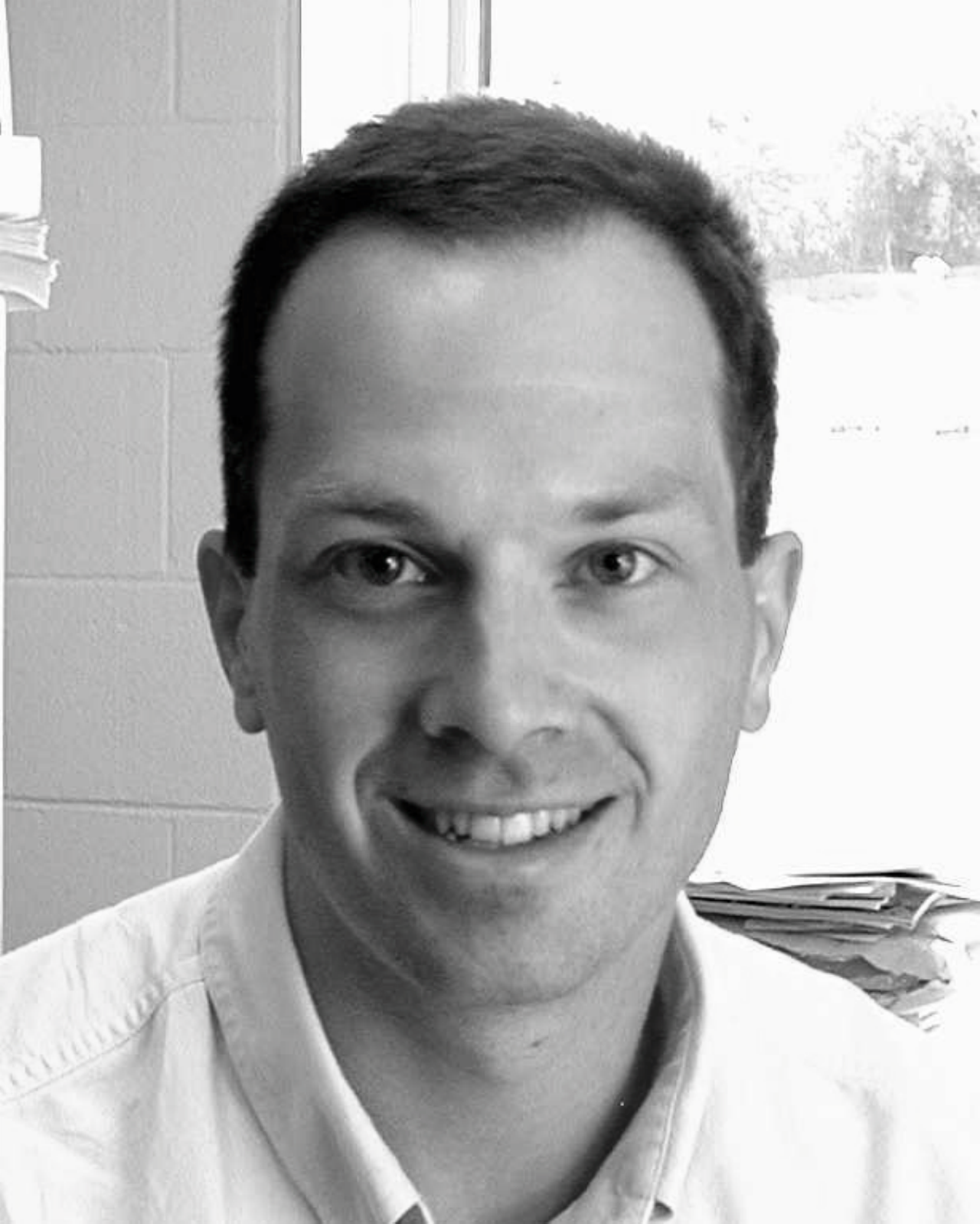}}]{Fran{\c c}ois Michaud}
(M'90) received his bachelor’s degree ('92), Master’s degree ('93) and Ph.D. degree ('96) in electrical engineering from the Universit\'e de Sherbrooke, Qu\'ebec Canada. 

After completing postdoctoral work at Brandeis University, Waltham MA ('97), he became a faculty member in the Department of Electrical Engineering and Computer Engineering of the Universit\'e de
Sherbrooke, and founded IntRoLab, a research laboratory working on designing intelligent autonomous systems that can assist humans in living environments. His research interests are in architectural methodologies for intelligent decision-making and design of interactive autonomous mobile robots. 

Prof. Michaud held a Canada Research Chair (2001-11) in Mobile Robots and Autonomous Intelligent Systems, and is the Director of the Interdisciplinary Institute for Technological Innovation (3IT). He is a member of IEEE, AAAI and OIQ (Ordre des ing\'enieurs du Qu\'ebec).
\end{IEEEbiography}

\end{document}